\documentclass[letterpaper]{article} 
\usepackage{aaai25}  
\usepackage{times}  
\usepackage{helvet}  
\usepackage{courier}  
\usepackage[hyphens]{url}  
\usepackage{graphicx} 
\usepackage{natbib}  
\usepackage{caption} 
\usepackage{algorithm}
\usepackage{algorithmic}
\usepackage{amssymb}
\usepackage[none]{hyphenat}
\usepackage{multirow}
\usepackage{color}
\usepackage{float}
\usepackage{amsmath}
\usepackage[table]{xcolor}

\usepackage{newfloat}
\usepackage{listings}
\DeclareCaptionStyle{ruled}{labelfont=normalfont,labelsep=colon,strut=off} 
\floatstyle{ruled}
\newfloat{listing}{tb}{lst}{}
\floatname{listing}{Listing}
\title{Accessible, At-Home Detection of Parkinson’s Disease\\ via Multi-Task Video Analysis}
\author{
    Md Saiful Islam\textsuperscript{\rm 1}\thanks{Corresponding author (email: mislam6@ur.rochester.edu).}, Tariq Adnan\textsuperscript{\rm 1}, Jan Freyberg\textsuperscript{\rm 2}, Sangwu Lee\textsuperscript{\rm 1}, Abdelrahman Abdelkader\textsuperscript{\rm 1},\\
    Meghan Pawlik\textsuperscript{\rm 3}, Cathe Schwartz\textsuperscript{\rm 4}, Karen Jaffe\textsuperscript{\rm 4}, Ruth B. Schneider\textsuperscript{\rm 3}, Ray Dorsey\textsuperscript{\rm 3}, Ehsan Hoque\textsuperscript{\rm 1}
}
\affiliations{
    \textsuperscript{\rm 1}University of Rochester, New York, United States\\
    \textsuperscript{\rm 2}Google Research, London, United Kingdom\\
    \textsuperscript{\rm 3} University of Rochester Medical Center, New York, United States\\
    \textsuperscript{\rm 4} InMotion, Ohio, United States\\

}

\usepackage{bibentry}

\begin{document}

\maketitle

\begin{abstract}
Limited accessibility to neurological care leads to under-diagnosed Parkinson's Disease (PD), preventing early intervention. Existing AI-based PD detection methods primarily focus on unimodal analysis of motor or speech tasks, overlooking the multifaceted nature of the disease. To address this, we introduce a large-scale, multi-task video dataset of $1102$ sessions (each containing videos of finger tapping, facial expression, and speech tasks captured via webcam) from $845$ participants ($272$ with PD). We propose a novel Uncertainty-calibrated Fusion Network (\emph{UFNet}) that leverages this multimodal data to enhance diagnostic accuracy. \emph{UFNet} employs independent task-specific networks, trained with Monte Carlo Dropout for uncertainty quantification, followed by self-attended fusion of features, with attention weights dynamically adjusted based on task-specific uncertainties. We randomly split the participants into training (60\%), validation (20\%), and test (20\%) sets to ensure patient-centered evaluation. \emph{UFNet} significantly outperformed single-task models in terms of accuracy, area under the ROC curve (AUROC), and sensitivity while maintaining non-inferior specificity. Withholding uncertain predictions further boosted the performance, achieving $88.0 \pm 0.3\%$ accuracy, $93.0 \pm 0.2\%$ AUROC, $79.3 \pm 0.9\%$ sensitivity, and $92.6 \pm 0.3\%$ specificity, at the expense of not being able to predict for $2.3 \pm 0.3\%$ data ($\pm$ denotes 95\% confidence interval). Further analysis suggests that the trained model does not exhibit any detectable bias across sex and ethnic subgroups and is most effective for individuals aged between 50 and 80. By merely requiring a webcam and microphone, our approach facilitates accessible home-based PD screening, especially in regions with limited healthcare resources.
\end{abstract}

%
\small
\begin{links}
    \link{Code \& Dataset}{https://github.com/ROC-HCI}
    \link{Published version}{https://doi.org/10.1609/aaai.v39i27.35031}
    \link{Demo}{https://parktest.net/demo}
\end{links}
\normalsize

\newpage

\section{Introduction}
Due to limited access to neurological care, many individuals, particularly in underserved regions~\cite{kissani2022does}, live with Parkinson's disease---the fastest growing neurological disorder~\cite{dorsey2018emerging}---without even knowing it. In many instances, when diagnosed late with this incurable disease, the condition could have already progressed significantly, limiting the usefulness of available medications. Imagine a future where individuals can remotely assess their risk for PD by simply visiting a website, activating their webcam and microphone, and completing a series of standardized tasks. This accessible approach could empower people to seek early intervention and treatment, potentially improving their quality of life.

Detecting PD is particularly challenging due to the variability of individual symptoms. For example, while vocal impairment is a common PD symptom~\cite{ho1999speech}, patients may also exhibit PD through other modalities, such as facial expression (e.g., hypomimia~\cite{gunnery2016relationship}) or motor function (e.g., bradykinesia~\cite{bologna2023redefining}). Consequently, models relying solely on a single task may yield suboptimal performance. To address this, we introduce a large video dataset featuring webcam recordings of individuals performing three tasks: (i) finger-tapping (motor function), (ii) smiling (facial expression), and (iii) uttering a pangram\footnote{pangram: sentence containing all the letters of the alphabet.} (speech). The dataset is collected from 845 unique participants (272 with PD) from diverse demographics who recorded all of these tasks successfully (about 20\% of them recorded multiple times), resulting in 1102 videos for each task and a total of 3306 videos. To our knowledge, this is the first multi-task video dataset for PD screening. 

We propose a novel two-stage classification model to distinguish between individuals with and without Parkinson's disease (PD) using our dataset. First, each task is independently modeled with neural networks trained using Monte Carlo dropout (MC dropout) to generate predictions and associated uncertainties. Next, we introduce the \textbf{U}ncertainty-calibrated \textbf{F}usion \textbf{N}etwork (\emph{UFNet}), which aggregates features from multiple tasks through an attention mechanism~\cite{vaswani2017attention}, while calibrating the attention scores based on task-specific uncertainties. \emph{UFNet} is designed to produce PD/Non-PD predictions and uses MC dropout during training and inference to quantify prediction confidence. By filtering out low-confidence predictions, it increases patient safety. Our evaluations on a subject-separated test set demonstrate that \emph{UFNet} significantly outperforms single-task models and multi-modal fusion baselines. The model is computationally efficient, employing only a few shallow neural networks and a single self-attention module, making it suitable for deployment on smartphones or personal computers.

Here is a summary of our key contributions:
\begin{itemize}
    \item We introduce the first-ever large-scale, multi-task video dataset for Parkinson's disease screening collected from 845 individuals with diverse demography. Although we cannot share the raw videos to protect identifiable patient information, we release the dataset as extracted features.

    \item We propose a novel, effective multimodal fusion model named \emph{UFNet}, achieving $88.0 \pm 0.3\%$ accuracy and $93.0 \pm 0.2\%$ AUROC on a subject-separated test dataset. The proposed model significantly outperformed single-task models and multimodal fusion baselines.
\end{itemize}

\section{Related Works}
Traditionally, Parkinson's Disease (PD) is diagnosed by a clinician based on medical history and a clinical examination, typically including completing standardized tasks and rating each task according to the MDS-UPDRS guidance~\cite{goetz2008movement}. Recently, a cerebrospinal fluid (CSF) based $\alpha$-synuclein seed amplification assay has been developed~\cite{siderowf2023assessment}, offering a potential diagnostic biomarker. However, access to clinical care can be limited, and diagnostic methods relying on the collection of CSF are invasive, costly, and burdensome to the patients.

Recent research has leveraged machine learning (ML) and sensors to enable remote assessment of PD. For example, breathing signals obtained from reflected radio waves can detect PD accurately when analyzed by an ML model~\cite{yang2022artificial}. Additionally, body-worn sensors have been successfully used to monitor clinical features such as dyskinesia and gait disturbances associated with PD~\cite{moreau2023overview}. However, wearable sensors face challenges related to cost, comfort, and ease of use, which limit the scalability of these techniques for global adoption.

Advanced pose tracking models like MediaPipe~\cite{lugaresi2019mediapipe} enable extracting precise clinical features from recorded videos, later used for PD screening~\cite{jin2020diagnosing} or progression tracking~\cite{islam2023using}. However, existing methods for video analysis suffer from major limitations -- small cohort size~\cite{jin2020diagnosing}, and reliance on single modality~\cite{tsanas2009accurate, ali2020spatio, ali2021analyzing, rahman2021detecting, adnan2023unmasking}. Symptoms of PD are multi-faceted and may affect individuals differently. For instance, one individual may face speech difficulty but retain normal motor functionality, while others may have prominent hypomimia (i.e., reduced facial expression) or bradykinesia (i.e., slowness of movement). Therefore, PD detection models may need to consider multiple modalities for improved efficacy.

This work addresses a critical gap in the field, which has persisted due to the lack of multimodal, diverse, and naturalistic data. By utilizing the PARK framework~\cite{langevin2019park} and inter-disciplinary collaboration, we have collected videos of multiple tasks from a large, diverse cohort, enabling us to develop an accurate PD screening model. We hope the community will engage with this dataset and continue to advance the model.

\section{Dataset}

\subsection{Standardized Tasks} We select three standardized tasks that can be easily completed using a computer webcam and microphone, with or without external supervision: 

\textbf{Finger-tapping:} Participants tap their thumb with the index finger ten times as fast as possible, first with the right hand, then with the left. The finger-tapping task is completed following the MDS-UPDRS scale to measure bradykinesia in the upper limb, a key sign of PD~\cite{hughes1992accuracy}.

\textbf{Smile:} Participants mimic a smile expression three times, alternating with a neutral face. Studies suggest the task captures signs of hypomimia~\cite{bandini2017analysis, adnan2023unmasking}, even though the expression may be unnatural.

\textbf{Speech:} Participants utter a script, ``\textit{The quick brown fox jumps over a lazy dog. The dog wakes up and follows the fox into the forest. But again, the quick brown fox jumps over the lazy dog.}'' The first sentence is a pangram, containing all the letters of the English alphabet, and the later sentences are added to obtain a longer speech segment. Prior research identified this task as a promising way of screening PD~\cite{rahman2021detecting}.

\subsection{Participant Recruitment} 
We recruited participants (with and without PD) from a diverse background (see Table \ref{tab:demographic_infromation}) through multiple channels, including a brain health study registry, social media, a PD wellness center, and clinician referrals. Among the approximately 1,400 individuals who recorded at least one task, 845 completed all three standardized tasks (comprising this dataset). Participants are primarily from the US, with diverse demographic characteristics: 52.7\% female, a mean age of 61.9 years (22.5\% over 70), and representation from various ethnic backgrounds. Among the 272 (32.2\%) participants who had PD, 233 had their diagnosis confirmed through clinical evaluation, while 39 self-reported their condition. This study was approved by the institutional review board (IRB) of the University of Rochester.

\begin{table*}[!htb]
    \centering
    \resizebox{0.8\linewidth}{!}{
    \begin{tabular}{llccc}
\textbf{Subgroup} & \textbf{Attribute} & \textbf{With PD} & \textbf{Without PD} & \textbf{Total} \\ \hline
 & Number of participants & 272 & 573 & 845 \\ \hline
Sex, n(\%) & Female & 122 (44.9\%) & 323 (56.4 \%) & 445 (52.7\%) \\ 
  & Male & 147 (54.0\%) & 250 (43.6 \%) & 397 (47.0\%) \\ 
  & Nonbinary & 1 (0.4\%) & 0 (0.0 \%) & 1 (0.1\%) \\
  & Unknown & 2 (0.7\%) & 0 (0.0 \%) & 2 (0.2\%) \\ 
\hline
\begin{tabular}[c]{@{}l@{}}Age in years, n (\%)\\ (range: 18.0 - 93.0, mean: 61.9)\end{tabular}
 & Below 20 & 0 (0.0 \%) & 6 (1.0 \%) & 6 (0.7 \%) \\
 & 20-29 & 1 (0.4 \%) & 28 (4.9 \%) & 29 (3.4 \%) \\
 & 30-39 & 2 (0.7 \%) & 19 (3.3 \%) & 21 (2.5 \%) \\
 & 40-49 & 6 (2.2 \%) & 17 (3.0 \%) & 23 (2.7 \%) \\
 & 50-59 & 33 (12.1 \%) & 119 (20.8 \%) & 152 (18.0 \%) \\
 & 60-69 & 94 (34.6 \%) & 231 (40.3 \%) & 325 (38.5 \%) \\
 & 70-79 & 98 (36.0 \%) & 76 (13.3 \%) & 174 (20.6 \%) \\
 & 80 and above & 12 (4.4 \%) & 4 (0.7 \%) & 16 (1.9 \%) \\
 & Unknown & 26 (9.6 \%) & 73 (12.7 \%) & 99 (11.7 \%) \\
\hline 
Ethnicity, n (\%)
 & American Indian or Alaska Native & 1 (0.4 \%) & 0 (0.0 \%) & 1 (0.1\%) \\
 & Asian & 3 (1.1 \%) & 34 (5.9 \%) & 37 (4.4\%) \\
 & Black or African American & 3 (1.1 \%) & 29 (5.1 \%) & 32 (3.8\%) \\
 & white & 163 (59.9 \%) & 463 (80.8 \%) & 626 (74.1\%) \\
 & Others & 2 (0.7 \%) & 3 (0.5 \%) & 5 (0.6\%) \\
 & Unknown & 100 (36.8 \%) & 44 (7.7 \%) & 144 (17.0\%) \\
\hline
\begin{tabular}[c]{@{}l@{}}Disease duration in years, n (\%)\\ (range: 1.0 - 24.0, mean: 6.6)\end{tabular}
& \textless{}=2 &19 (6.99\%) & - & - \\
&2-5 &36 (13.24\%)  & - & - \\
&5-10 &28 (10.29\%)  & - & - \\
&10-15 &10 (3.68\%)  & - & - \\
&15-20 &7 (2.57\%)  & - & - \\
& \textgreater{}20 &2 (0.74\%)  & - & - \\
& Unknown &170 (62.5\%)  & - & - \\
\hline 
Cohort, n (\%)
 & Home & 39 (14.3 \%) & 399 (69.6 \%) & 438 (51.8 \%) \\
 & Clinic & 91 (33.5 \%) & 107 (18.7 \%) & 198 (23.4 \%) \\
 & PD wellness center & 142 (52.2 \%) & 67 (11.7 \%) & 209 (24.7 \%) \\
\hline
\end{tabular}%
}
\caption{Summary of demographic information. Participants recruited via clinician referrals (Clinic cohort) and the PD wellness center were clinically diagnosed, but the other participants' diagnosis status was self-reported. Information related to sex, age, ethnicity, and disease duration were self-reported by the participants. It is important to note that demographic information was optional, and missing data is indicated as ``Unknown.''}
\label{tab:demographic_infromation}
\end{table*}

\setlength{\tabcolsep}{1mm} 
\begin{table}
\centering
\small
\begin{tabular}{clcc}
\multicolumn{2}{c}{{\textbf{Task/Split}}} & {\textbf{\#Sessions, PD \%}} & {\textbf{\#Participants, PD \%}} \\ \hline
\multicolumn{2}{l}{\textbf{Finger-tapping}}                    & \textbf{1374, 41.3\%}                                                                                       & \textbf{1167, 36.6\%}                                                                           \\
                          & Training                           & 945, 43.9\%                                                                                                & 819, 38.8\%                                                                                                                                                                         \\
                          & Validation                         & 221, 37.6\%                                                                                                 & 172, 32.6\%                                                                                                                                                                        \\
                          & Test                               & 208, 33.2\%                                                                                                 & 176, 30.1\%                                                                                                                                                                        \\ \hline
\multicolumn{2}{l}{\textbf{Smile}}                             & \textbf{1684, 32.8\%}                                                                                       & \textbf{1357, 28.5\%}                                                                                                                                                   \\
                          & Training                           & 1021, 33.2\%                                                                                                & 824, 28.4\%                                                                                                                                                           \\
                          & Validation                         & 342, 33.9\%                                                                                                 & 266, 28.6\%                                                                                                                                                           \\
                          & Test                               & 321, 30.5\%                                                                                                 & 267, 28.8\%                                                                                                                                                          \\ \hline
\multicolumn{2}{l}{\textbf{Speech}}                            & \textbf{1655, 33.9\%}                                                                                       & \textbf{1265, 28.9\%}                                                                                                                                               \\
                          & Training                           & 1007, 35.3\%                                                                                                 & 769, 29.0\%                                                                                                                                                           \\
                          & Validation                         & 338, 33.7\%                                                                                                 & 252, 29.0\%                                                                                                                                                            \\
                          & Test                               & 310, 29.7\%                                                                                                 & 244, 28.7\%                                                                                                                                                            \\ \hline
\multicolumn{2}{l}{\textbf{All tasks}}                         & $\mathbf{1102, 41.8\%}$                                                                                       & \textbf{845, 32.2\%}                                                                                                                                                      \\
                          & Training                           & 690, 45.1\%                                                                                                 & 516, 32.6\%                                                                                                                                                           \\
                          & Validation                         & 215, 38.1\%                                                                                                 & 167, 32.9\%                                                                                                                                                            \\
                          & Test                               & 197, 34.9\%                                                                                                 & 162, 30.2\%                                                                                                                                                             \\ \hline
\end{tabular}
\caption{Summary of dataset splits.}
\label{tab:dataset}
\end{table}

\subsection{Dataset Splits} 
Most participants completed all three tasks, but some did not, and some task videos were discarded due to feature extraction failures, often caused by inaccurate task completion. Instead of discarding videos of participants with missing tasks, we trained task-specific models with all available videos for each task. The multi-task model is then trained on the participants who completed all three tasks (Table \ref{tab:dataset}). We split the datasets based on the participants to ensure patient-centric evaluation. All participants were randomly assigned into training (60\%), validation (20\%), and test (20\%) sets, with stratification to maintain a similar ratio of individuals with and without PD across folds. Both task-specific and multi-task models are validated and tested on the same participant cohort, with no model seeing any data from these participants during training.

\section{Our Approach}
\begin{figure*}[t]
    \centering
    \includegraphics[width=0.90\linewidth]{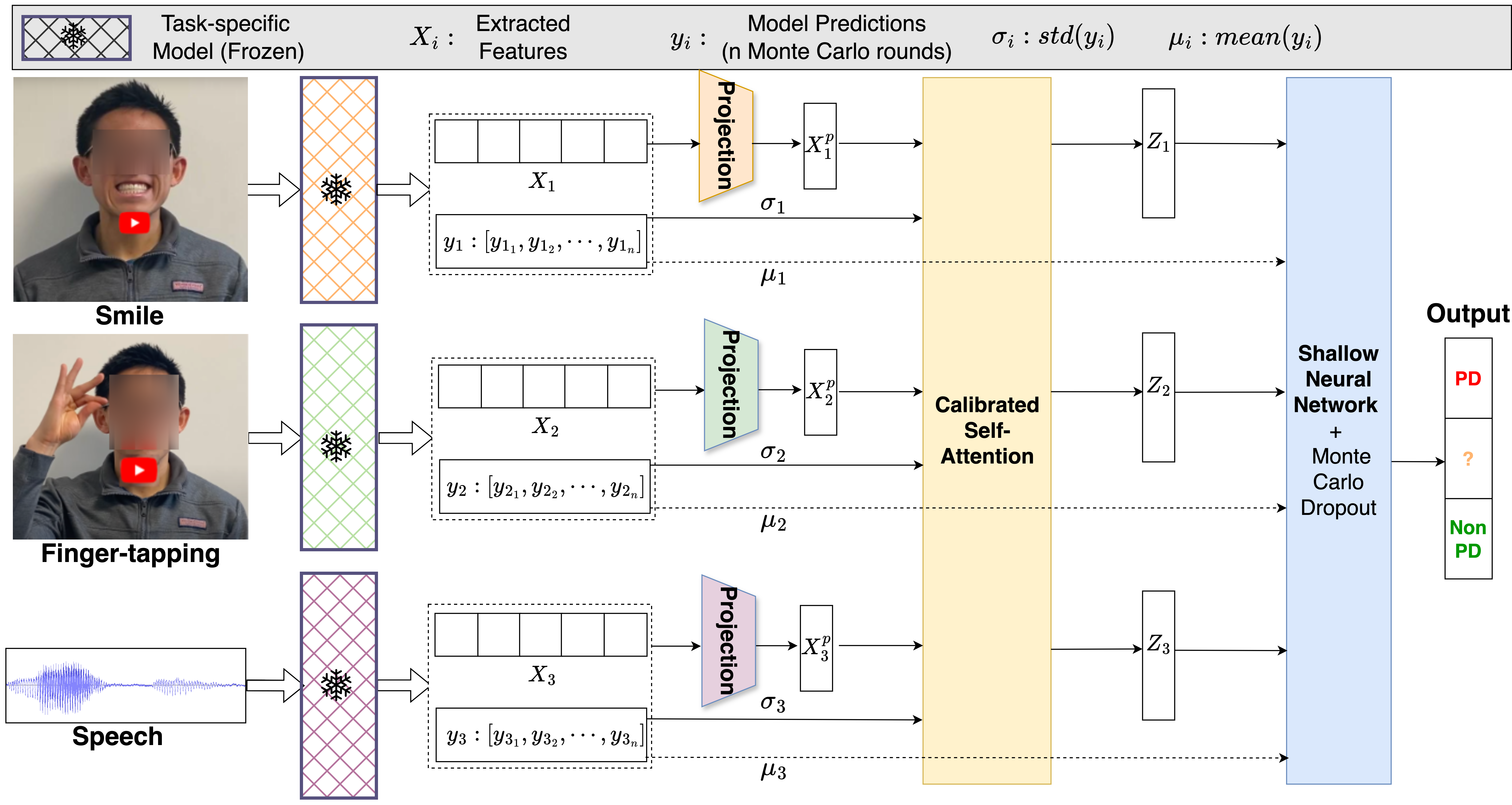}
    \caption{\textbf{An illustrative overview of the proposed model.} Task-specific models are shallow neural networks trained with task-specific features and MC dropout. Standard deviations in multiple rounds of inference obtained from the task-specific models are used to calibrate the attention scores when fusing task-specific features. Finally, another shallow neural network is trained to differentiate individuals with and without PD and withhold prediction when the model is uncertain.}
    \label{fig:method_overview}
\end{figure*}

\subsection{Feature Extraction}
We rely on prior literature to extract task-specific video features. For the smile and finger-tapping tasks, clinically meaningful hand-crafted features were the better choice against deep video models such as ViViT~\cite{arnab2021vivit}, TimeSformer~\cite{bertasius2021space}, VideoMAE~\cite{tong2022videomae}, and Uniformer~\cite{li2023uniformer}, based on their performance and suitability for low-resource deployment. WavLM~\cite{chen2022wavlm} has been identified as the most effective for speech-based PD screening by a recent work~\cite{adnan2024novel}.

\subsubsection{Finger-tapping features.} ~\citet{islam2023using} extracted 65 features to analyze the finger-tapping task for assessing PD severity. Using MediaPipe hand~\cite{grishchenko2022blazepose} to detect hand movements and key points, they measured clinically relevant features such as finger-tapping speed, amplitude, and interruptions. We apply this technique to both hands, extracting 130 features in total.

\subsubsection{Smile features.}
We used 42 facial features from smile videos extracted by ~\citet{adnan2023unmasking}. These features, extracted using OpenFace~\cite{baltruvsaitis2016openface} and MediaPipe, capture key PD markers such as eye blinking, lip separation, mouth opening, and intensity of facial muscle movements.

\subsubsection{Speech features.} We extracted 1024-dimensional embeddings from a pre-trained WavLM~\cite{chen2022wavlm} model to encode the speech task. WavLM, trained on massive amounts of speech data, excels at capturing acoustic characteristics of speech, making it useful for tasks like speaker identification, speech, and emotion recognition.

\subsection{Task-specific Models}
Each task utilizes a separate machine learning model to distinguish between individuals with and without PD. These models have three main components:

\begin{itemize}
    \item \textbf{Feature selection and scaling:} We calculate pairwise correlation among the features to drop highly correlated features. Feature values are then scaled using \emph{StandardScaler} or \emph{MinMaxScaler}. Whether to apply feature selection or scaling, the correlation threshold and the scaling method are hyper-parameters tuned on the validation set.

    \item \textbf{Shallow neural networks:} The network consists of 1-2 linear layers, with a single output neuron in the last layer. We use \texttt{sigmoid} activation with the output layer and \texttt{ReLU} with the other layers to add non-linearity.

    \item \textbf{Monte Carlo dropout (MC dropout):} To improve model robustness and estimate prediction uncertainty, we employ MC dropout~\cite{atighehchian2019baal}. This technique allows using dropout for both training and inference. Therefore, we can obtain predictions from multiple rounds during testing and the mean ($\mu$) and standard deviation ($\sigma$) of these predictions. We expect a relatively higher standard deviation across these predictions when the model is uncertain. Therefore, $\sigma$ is used to estimate model uncertainty.
\end{itemize}

The task-specific models are trained using binary cross-entropy loss and either SGD or AdamW optimizer.

\subsection{Uncertainty-calibrated Fusion Network, \emph{UFNet}}
The fully trained task-specific models remain frozen during the training of \emph{UFNet}. For each task $i$, the extracted features ($X_i \in \mathbb{R}^{d_{X_i}}$), predicted mean probability ($\mu_i$), and uncertainty in the prediction ($\sigma_i$) are input to \emph{UFNet}. The model then combines information from all the tasks through a series of steps to generate a final, more robust detection of PD (see Figure \ref{fig:method_overview} for an overview).

\subsubsection{Projection.} Since the size of the features may vary from task to task, they are first projected to the same dimension ($d$) using a projection layer. Each projection layer consists of a linear layer ($\mathbb{R}^{d_{X_i}} \rightarrow \mathbb{R}^{d}$) with MC dropout, followed by non-linear activation (\texttt{ReLU}) and layer normalization.

\subsubsection{Calibrated self-attention.} 
We employ self-attention to integrate the projected features ($X^p_i \in \mathbb{R}^d$) extracted from three distinct tasks. Unlike standard self-attention applications, our input sequence comprises task-specific feature vectors (sequence length = 3) rather than tokens of the same modality. For each task, projected features are transformed into query ($Q$), key ($K$), and value ($V$) vectors using linear projections. Attention weights are computed based on the similarity between queries and keys. 

To prioritize informative tasks, we adjust the attention scores to down-weight contributions from tasks with higher prediction uncertainty. Specifically, in addition to using traditional query-key similarity, we update the attention matrix $A$ with task-specific uncertainty scores $\Sigma = [\sigma_1, \sigma_2, \sigma_3]$ as, $$A = softmax(\frac{Q.K^T}{\sqrt{d}} - \eta \Sigma)$$ Hyper-parameter $\eta$ controls the degree of uncertainty-based down-weighting. After computing the attention scores, $X^p_i$ is converted into a contextualized representation $Z_i = \sum_{j=1}^{N} A_{i,j} V_j$, where $N = 3$ is the number of tasks.

\subsubsection{Shallow neural network.}
The contextualized representations ($Z_1, Z_2, Z_3$) obtained after self-attention are concatenated along with the task-specific (mean) predicted probabilities ($\mu_1, \mu_2, \mu_3$). The merged vector is then input to a shallow neural network similar to the one used for task-specific training. The network is trained with 30 rounds of MC dropout, and the average output is used as the final prediction (PD if the average output is more than 0.50, non-PD otherwise).

\subsubsection{Withholding predictions.}
Our proposed approach is designed for neurological screening, specifically to suggest clinical evaluation for individuals if the model identifies PD symptoms. Since mispredictions in this context could lead to significant consequences (e.g., unnecessary anxiety or delayed care), the model is designed to withhold predictions it identifies as uncertain. To achieve this, we implement two different techniques:
\begin{itemize}
    \item \textbf{MC Dropout:} For new test data, we generate $n$ predictions by running multiple rounds of Monte Carlo (MC) dropout. We compute the $95\%$ confidence interval (CI) for these predictions and classify them as uncertain if the CI contains the decision threshold (i.e., 0.50).
    \item \textbf{Conformal Prediction:} We partition the training data into a training set (80\%) and a calibration set (20\%). We calculate a conformity threshold using the calibration set and identify uncertain predictions when the prediction set contains both PD and Non-PD classes.
\end{itemize}
We also experiment with performing label smoothing~\cite{muller2019does} (which redistributes probability mass to reduce overconfidence) and/or Platt scaling~\cite{platt1999probabilistic} (which fits a logistic regression model to improve calibration) with conformal prediction~\cite{saunders1999transduction} to see whether they further improve model performance.

\subsubsection{Hyper-parameter tuning.}
We utilized Weights \& Biases\footnote{https://wandb.ai} to employ Bayesian hyper-parameter tuning within the defined search space (see Appendix B and C) to maximize the AUROC on the validation set. Finally, the \emph{UFNet} model is trained with binary cross-entropy loss and \texttt{SGD} optimizer with learning rate $0.0207$ and momentum $0.6898$. We selected 64 as the dimension for the query, key, and value vectors. The chosen value of dropout probability was $0.4960$, and $\eta$ was 81.8179. Since some hyper-parameters had continuous search space (e.g., the learning rate was tuned within $1e^{-5}$ and 1.0), we report the precise values of the selected hyper-parameters.

\section{Experiment Setup}
Experiments are conducted on an AMD Threadripper 3970x 32-core CPU with 256 GB RAM and two NVIDIA A6000 GPUs with 48 GB VRAM. We use Python (PyTorch deep learning framework) and Linux operating system. 

\subsection{Multimodal Baselines}
We compare our proposed \emph{UFNet} with four popular multimodal fusion approaches.

\textbf{Majority Voting.} The three task-specific models' predictions are combined to generate a single prediction. The final prediction is the class (PD or Non-PD) agreed upon by the majority (i.e., two or more). 

\textbf{Neural Late Fusion.} The logit scores from the task-specific models are input to a shallow neural network trained to predict a binary class, similar to logistic regression or other ensembling methods.

\textbf{Early Fusion Baseline.} Features from all three tasks are concatenated and input to a shallow neural network.

\textbf{Hybrid Fusion Baseline.} Task-specific features and prediction scores (logits) are provided as input to a shallow neural network. This network leverages both the input features and prediction scores, combining the strengths of early and late fusion approaches.

\subsection{Performance Reporting}
After hyper-parameter tuning, we train the selected model with 30 random seeds and evaluate it on the test set. We report the average and 95\% confidence interval (CI) of each performance metric. In addition to standard binary classification metrics, we report coverage (\% of cases where the model is confident enough to make a prediction) when appropriate. Furthermore, ensuring that the class probabilities output by a model approximates the actual probability is critical for patient safety. Therefore, we evaluate model calibration using expected calibration error (ECE)~\cite{nixon2019measuring} and Brier score~\cite{rufibach2010use}. Unless specified otherwise, a significant difference is identified by having no overlap in the 95\% CI, while non-inferiority indicates overlap.

\section{Results}

\subsection{Task-specific Model Performance}
\setlength{\tabcolsep}{1mm}
\begin{table}[!htb]
\centering
\small
\begin{tabular}{llccccc}
\multicolumn{2}{l}{\textbf{Task}}           & \textbf{Accuracy}        & \textbf{$F_1$ score}                                                        & \textbf{AUROC}\\ 
\hline
\multicolumn{2}{l}{\textbf{Finger-tapping}} &                                                                             &                                                                             &\\

            & w/o MC dropout            & $72.0 \pm 0.9$          & $60.2 \pm 1.3$          & $73.9 \pm 0.9$\\
            
            & w/ MC dropout               & $73.1 \pm 0.7$          & $61.7 \pm 0.9$          & $74.9 \pm 0.7$\\ 
            
            \hline
            
\multicolumn{2}{l}{\textbf{Smile}}          &                                                                             &                                                                             &                                                                             &                                                                             &\\
            & w/o MC dropout            & $75.6 \pm 0.2$          & $64.3 \pm 0.3$          & $83.2 \pm 0.1$\\
            
            & w/ MC dropout               & \underline{$77.6 \pm 0.2$} & \underline{$67.5 \pm 0.3$} & \underline{$83.6 \pm 0.1$}\\ 
            
            \hline
            
\multicolumn{2}{l}{\textbf{Speech}}         &                                                                             &                                                                             &\\

            & w/o MC dropout            & $84.5 \pm 0.3$          & $71.7 \pm 0.4$ & \underline{$\mathbf{89.4 \pm 0.2}$}\\
            
            & w/ MC dropout               & \underline{$\mathbf{85.1 \pm 0.2}$} & $\mathbf{72.1 \pm 0.6}$          & $87.8 \pm 0.1$\\ \hline
\end{tabular}
\caption{Performance of task-specific models with (w/) and without (w/o) MC dropout (in percentage). \underline{Underlined} metrics denote significantly better performance (within the same task), while \textbf{bold} metrics show overall best performance.}
\label{tab:mcdropout}
\end{table}

Among the three standardized tasks, speech is the most accurate for PD classification, and finger-tapping is the least accurate. The shallow neural network achieved $84.5 \pm 0.3\%$ accuracy and $89.4 \pm 0.2\%$ AUROC using only speech features. Models trained on right-hand tapping features performed better than those trained on left-hand features, indicating potential under-representation of left-handed individuals. However, the model trained on concatenated features from both hands significantly improved the $F_1$ score while being non-inferior in other metrics (see Appendix A). The finger-tapping model refers to the one trained on both-hand features for the remaining analysis.

MC dropout significantly boosted the performance of the smile model in all metrics. For the speech model, while MC dropout improved accuracy and balanced accuracy, it decreased AUROC and AUPRC. The performance of the finger-tapping model was not notably affected. Due to the additional benefits of MC dropout in modeling prediction uncertainty, the multi-task models (\emph{UFNet} and other baselines) were trained with MC dropout unless specified otherwise. See Table \ref{tab:mcdropout} for details.

\subsection{Effect of multi-task combinations}
Combining multiple tasks using the proposed \emph{UFNet} model enhanced performance (Table \ref{tab:multi-tasks}). For instance, the AUROC scores for all multi-task models were significantly better than the corresponding single-task scores. Although the finger-tapping task alone was the weakest for detecting PD, its features may have complemented other task features, resulting in significant improvements in most metrics. Combining all three tasks significantly improved all reported metrics, achieving an accuracy of $87.3 \pm 0.3\%$, AUROC of $92.8 \pm 0.2\%$, and an $F_1$ score of $81.0 \pm 0.6\%$.

\setlength{\tabcolsep}{1mm}
\begin{table}[]
\centering
\small
\begin{tabular}{lccc}
\hline
\textbf{Task Combination}                & \textbf{Accuracy}                                                                    & $\mathbf{F_1}$ \textbf{score}                                                                 & \textbf{AUROC}\\ \hline
All three tasks
& $\mathbf{87.3 \pm 0.4}$          & $\mathbf{81.0 \pm 0.6}$ & $\mathbf{92.8 \pm 0.2}$\\\hline 

Tapping + Smile          & $78.0 \pm 0.8$          & $65.6 \pm 1.7$          & $84.8 \pm 0.5$\\ 
Tapping + Speech         & $84.1 \pm 0.3$          & $77.3 \pm 0.4$          & $91.4 \pm 0.2$\\ 
Smile + Speech                  & $85.2 \pm 0.3$ & $75.0 \pm 0.4$          & $91.2 \pm 0.1$\\\hline

Tapping                  & $73.1 \pm 0.7$ & $61.7 \pm 0.9$          & $74.9 \pm 0.7$\\
Smile                  & $77.6 \pm 0.2$ & $67.5 \pm 0.3$          & $83.6 \pm 0.1$\\
Speech                  & $85.1 \pm 0.2$ & $72.1 \pm 0.6$          & $87.8 \pm 0.1$\\\hline
\end{tabular}
\caption{Performance of models (as percentages) trained on different combinations of the standardized tasks. ``Tapping'' is used as the short form of the finger-tapping task. The \textbf{bold} metrics denote best performance.}
\label{tab:multi-tasks}
\end{table}

\subsection*{Comparison against baselines}

\setlength{\tabcolsep}{0.6mm}
\begin{table*}[t]
\centering
\small
\begin{tabular}{lccccccc}
\textbf{Model}                                                                                                                           & \textbf{Accuracy}                                                                 & \begin{tabular}[c]{@{}c@{}}\textbf{Balanced}\\ \textbf{Accuracy}\end{tabular}              & \multicolumn{1}{l}{\textbf{AUROC}}                                                & $\mathbf{F_1}$ \textbf{score}                                                              & \textbf{Precision}                & \begin{tabular}[c]{@{}c@{}}\textbf{Sensitivity}\\ \textbf{(Recall)}\end{tabular}           & \multicolumn{1}{l}{\textbf{Specificity}}\\ \hline
\textbf{Baseline models} & & & & & & &\\
Majority Voting                                                                                                                 & 85.3                                                                    & 83.9                                                                    & 89.6                                                                    & 78.2                                                                    & 80.0                                                                     & 76.5                                                                    & 89.9\\ 
Neural Late Fusion                                                                                                              & $84.1 \pm 0.4$          & $81.3 \pm 4.8$          & $91.7 \pm 2.2$          & $73.2 \pm 8.3$          & $73.5 \pm 7.5$          & $76.3 \pm 9.4$          & $88.2 \pm 2.7$\\ 
Early Fusion Baseline                                                                                                           & $83.6 \pm 0.6$          & $81.8 \pm 0.7$          & $91.0 \pm 0.2$          & $76.7 \pm 0.7$          & $75.4 \pm 1.1$          & $78.1 \pm 0.9$          & $86.5 \pm 0.8$\\ 
Hybrid Fusion Baseline                                                                                                          & $84.1 \pm 0.3$ & $82.4 \pm 0.4$ & $91.4 \pm 0.2$ & $77.3 \pm 0.4$ & $76.2 \pm 0.7$ & $\mathbf{78.6 \pm 0.6}$ & $87.0 \pm 0.6$\\ \hline
\textbf{Our model with different attention modules} & & & & & & &\\ 
Dot product self-attention~\cite{vaswani2017attention} & $85.5 \pm 0.4$ & $84.3 \pm 0.4$ & $\mathbf{92.9 \pm 0.2}$ & $78.3 \pm 0.6$ & $80.7 \pm 0.6$ & $76.1 \pm 1.1$ & $90.4 \pm 0.4$\\
LRFormer~\cite{ye2023mitigating} & \underline{$86.2 \pm 0.5$} & $85.1 \pm 0.5$ & $92.6 \pm 0.3$ & $79.5 \pm 0.7$ & $81.7 \pm 0.9$ & $77.6 \pm 1.0$ & $90.8 \pm 0.6$\\
UFNet (ours) & \underline{$\mathbf{87.3 \pm 0.4}$} & $\mathbf{86.4 \pm 0.4}$ & $92.8 \pm 0.2$ & $\mathbf{81.0 \pm 0.6}$ & \underline{$\mathbf{83.8 \pm 0.5}$} & $78.4 \pm 1.0$ & \underline{$\mathbf{92.0 \pm 0.3}$}\\
\hline
\textbf{Early vs. hybrid fusion in UFNet} & & & & & & &\\
Early Fusion & \underline{$86.7 \pm 0.5$} & $85.8 \pm 0.5$ & $92.7 \pm 0.3$ & $79.9 \pm 0.8$ & \underline{$83.3 \pm 0.7$} & $76.9 \pm 1.4$ & \underline{$91.9 \pm 0.4$}\\ 
Hybrid Fusion & \underline{$\mathbf{87.3 \pm 0.4}$} & $\mathbf{86.4 \pm 0.4}$ & $92.8 \pm 0.2$ & $\mathbf{81.0 \pm 0.6}$ & \underline{$\mathbf{83.8 \pm 0.5}$} & $78.4 \pm 1.0$ & \underline{$\mathbf{92.0 \pm 0.3}$}\\ 
\hline
\end{tabular}
\caption{Comparison of \emph{UFNet} performance (as percentages) against multimodal baselines and other attention modules. The \underline{underlined} metrics indicate significantly better performance than all four baselines. \textbf{Bold} metrics indicate the overall best performance across all choices. CI is not reported for majority voting since it does not involve any randomness.}
\label{tab:comparison}
\end{table*}

\setlength{\tabcolsep}{0.6mm}
\begin{table*}[t]
\centering
\small
\begin{tabular}{lccccccc}
\textbf{Model}                                                                                                                           & \textbf{Accuracy}                                                                 & \multicolumn{1}{l}{\textbf{AUROC}}                                                & $\mathbf{F_1}$ \textbf{score}                                                              & \textbf{Precision}                & \textbf{Recall}           \\ \hline
\textbf{Baseline models} & & & & &\\
Majority Voting                                                                                                                 & 85.3                                                                    & 89.6                                                                    & 78.2                                                                    & 80.0                                                                     & 76.5\\ 
Neural Late Fusion                                                                                                              & $84.1 \pm 0.4$          & $91.7 \pm 2.2$          & $73.2 \pm 8.3$          & $73.5 \pm 7.5$          & $76.3 \pm 9.4$\\ 
Early Fusion Baseline                                                                                                           & $83.6 \pm 0.6$    & $91.0 \pm 0.2$          & $76.7 \pm 0.7$          & $75.4 \pm 1.1$          & $78.1 \pm 0.9$\\ 
Hybrid Fusion Baseline                                                                                                          & $84.1 \pm 0.3$    & $91.4 \pm 0.2$ & $77.3 \pm 0.4$ & $76.2 \pm 0.7$ & $\mathbf{78.6 \pm 0.6}$\\ \hline
\textbf{Attention variants} & & & & & & &\\ 
$\text{Dot product self-attention}^{11}$ & $85.5 \pm 0.4$ & $\mathbf{92.9 \pm 0.2}$ & $78.3 \pm 0.6$ & $80.7 \pm 0.6$ & $76.1 \pm 1.1$\\
$\text{LRFormer}^{12}$ & \underline{$86.2 \pm 0.5$} & $92.6 \pm 0.3$ & $79.5 \pm 0.7$ & $81.7 \pm 0.9$ & $77.6 \pm 1.0$\\
UFNet (ours) & \underline{$\mathbf{87.3 \pm 0.4}$} & $92.8 \pm 0.2$ & $\mathbf{81.0 \pm 0.6}$ & \underline{$\mathbf{83.8 \pm 0.5}$} & $78.4 \pm 1.0$\\
\hline
\textbf{Early vs. hybrid fusion} & & & & & & &\\
Early Fusion & \underline{$86.7 \pm 0.5$} & $92.7 \pm 0.3$ & $79.9 \pm 0.8$ & \underline{$83.3 \pm 0.7$} & $76.9 \pm 1.4$\\ 
Hybrid Fusion & \underline{$\mathbf{87.3 \pm 0.4}$} & $92.8 \pm 0.2$ & $\mathbf{81.0 \pm 0.6}$ & \underline{$\mathbf{83.8 \pm 0.5}$} & $78.4 \pm 1.0$\\ 
\hline
\end{tabular}
\caption{Comparison of \emph{UFNet} performance (as percentages) against multimodal baselines and other attention modules. The \underline{underlined} metrics indicate significantly better performance than all four baselines. \textbf{Bold} metrics indicate the overall best performance across all choices. CI is not reported for majority voting since it does not involve any randomness.}
\label{tab:comparison2}
\end{table*}

The proposed uncertainty-calibrated fusion network (\emph{UFNet}) outperformed the baseline methods in most metrics. More specifically, \emph{UFNet} significantly improved PD classification accuracy, precision, and specificity compared to any of the four baselines. It also achieved higher balanced accuracy, AUROC, and $F_1$ score while maintaining statistically non-inferior sensitivity (Table \ref{tab:comparison}).

\subsection{Ablation results}
To investigate the significance of our introduced uncertainty-calibrated self-attention module, we replace it with the standard dot product self-attention~\cite{vaswani2017attention} and LRFormer~\cite{ye2023mitigating} -- a recently proposed self-attention module that prevents transformer overconfidence via Lipschitz Regularization. In addition, \emph{UFNet} uses task-specific features and prediction scores, making it a hybrid fusion approach. We implement an early fusion approach to analyze the effect of removing task-specific predictions from \emph{UFNet}. 

The calibrated self-attention module outperformed the other two attention modules in all metrics except AUROC. While the dot product self-attention yielded 0.1\% higher (non-significant) AUROC, \emph{UFNet} was significantly better in most other metrics. LRFormer performed closely in most metrics. Finally, including task-specific predictions as additional input (hybrid fusion) enhanced performance over the early fusion approach. Since our task-specific models were trained with more subjects, they likely better understand task-specific predictions, and incorporating them improved final performance. 

\subsection{Withholding uncertain predictions}
The trivial usage of MC dropout improved prediction accuracy. With dropped uncertain predictions, the best \emph{UFNet} (hybrid fusion) model achieved $88.0 \pm 0.3\%$ accuracy, $87.1 \pm 0.3\%$ balanced accuracy, $93.0 \pm 0.2\%$ AUROC, and $81.8 \pm 0.5\%$ $F_1$ score -- a clear improvement over our base model. It also achieved better precision, sensitivity, and specificity. However, the model could now predict $97.8 \pm 0.3\%$ of the data where it is confident enough. The model's expected calibration error (ECE) was $5.4 \pm 0.5\%$, and the Brier score was $0.097 \pm 0.002$, indicating that the model-predicted probabilities aligned well with actual disease probabilities.

\begin{table*}[]
\centering
\begin{tabular}{lcccccc}
\multicolumn{1}{c}{\textbf{Withholding Approach}}                                                 & \multicolumn{1}{c}{\textbf{Accuracy}} & \multicolumn{1}{c}{\textbf{Precision}} & \textbf{Recall} & \textbf{F1 Score} & \textbf{AUROC} & \textbf{Coverage} \\ \hline
MC Dropout + 95\% CI                                                                              & $88.0 \pm 0.3$                        & $84.6 \pm 0.5$                         & $79.3 \pm 0.9$  & $81.8 \pm 0.6$    & $93.0 \pm 0.2$ & \underline{$\mathbf{97.8 \pm 0.3}$}    \\
Conformal Prediction                                                                              & $93.9 \pm 0.7$                        & $87.0 \pm 1.4$                         & $89.0 \pm 2.5$  & $87.9 \pm 1.8$    & $93.9 \pm 0.7$ & $55.1 \pm 2.5$    \\
Label Smoothing + Conformal Prediction                                                            & $92.3 \pm 0.7$                                     & $84.9 \pm 1.2$                                      & $90.0 \pm 1.5$               & $87.3 \pm 1.0$                 & $94.4 \pm 0.6$              & $56.9 \pm 2.4$                 \\
Platt Scaling + Conformal Prediction                                                              & $91.9 \pm 0.8$                                     & $85.3 \pm 0.9$                                      & \underline{$\mathbf{97.6 \pm 0.7}$}               & $\mathbf{91.0 \pm 0.4}$                 & $92.8 \pm 2.2$              & $48.7 \pm 3.3$                 \\
\begin{tabular}[c]{@{}l@{}}Label Smoothing + \\ Platt Scaling + Conformal Prediction\end{tabular} & $\mathbf{94.2 \pm 0.5}$                                     & $\mathbf{87.4 \pm 1.3}$                                      & $92.9 \pm 1.3$               & $89.9 \pm 0.7$                 & $\mathbf{95.5 \pm 0.7}$              & $52.0 \pm 2.6$                 \\ \hline
\end{tabular}
\caption{\textbf{Different methods for withholding predictions.}}
\label{tab:ablation_withholding}
\end{table*}

Applying conformal prediction further reduced misprediction at the cost of a more frequent prediction withholding. Simultaneously applying label smoothing, Platt scaling, and conformal prediction resulted in a notable improvement in all metrics, resulting in $94.2 \pm 0.0.5\%$ accuracy, $89.9 \pm 0.7\%$ $F_1$ score, and $95.5 \pm 0.7\%$ AUROC. But this model could predict with enough certainty in only $52.0 \pm 2.6\%$ cases, throwing away approximately half of the test data. Please see Table \ref{tab:ablation_withholding} for details.

\subsection{Performance across demographic subgroups} 
\begin{figure}[t]
\centering
\includegraphics[width=\linewidth]{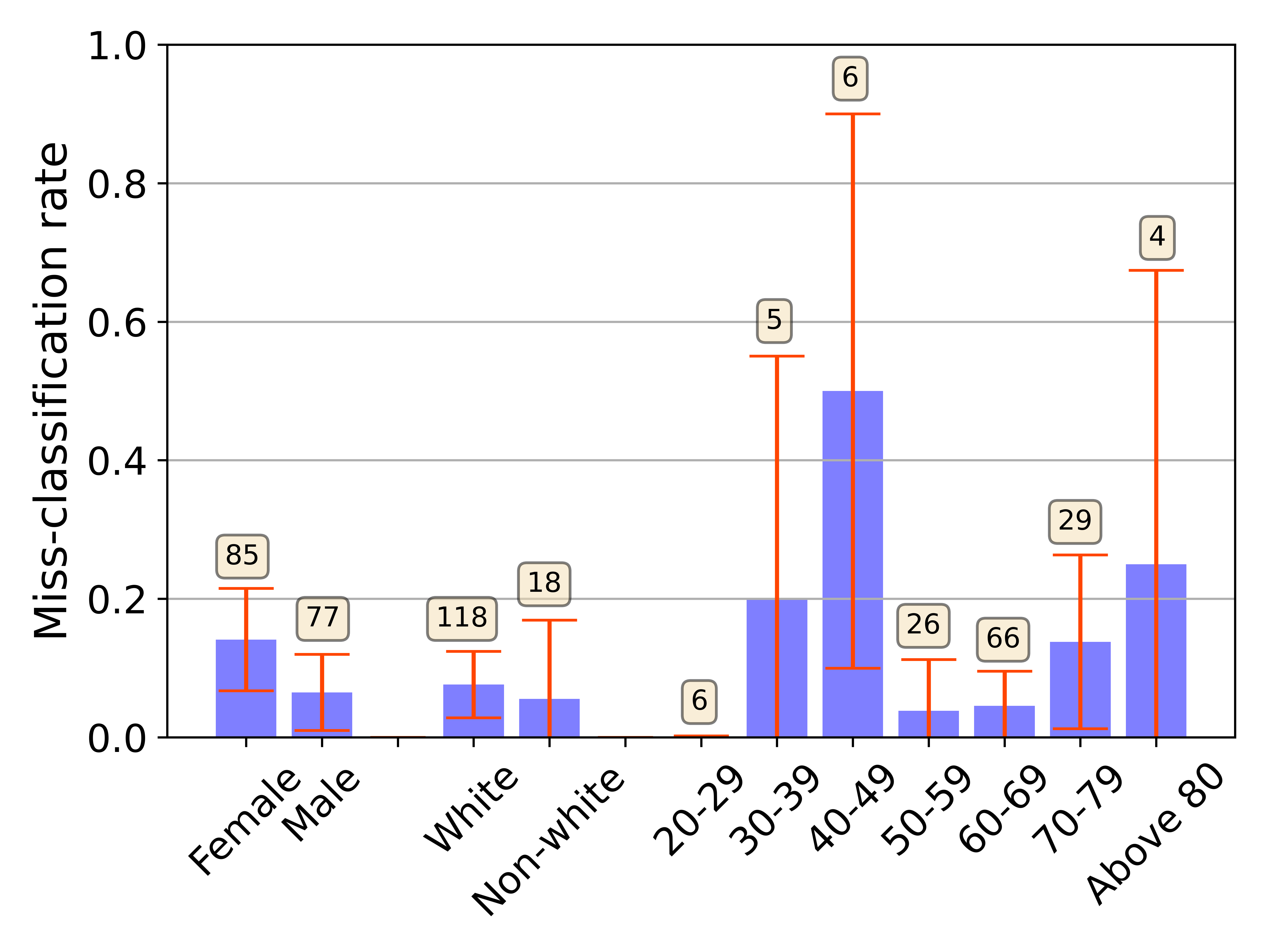}
\caption{Misclassification rate of the best \emph{UFNet} model across demographic subgroups. The analysis is done on the test set participants. Error bars demonstrate 95\% CIs.}
\label{fig:bias}
\end{figure}

No significant bias in model performance was observed (see Figure \ref{fig:bias}) based on sex or race in the test set (162 individuals). The average error (i.e., misclassification) rate across female participants ($n = 85$) was $14.1 \pm 7.4\%$, compared to $6.5 \pm 5.5\%$ for male participants ($n = 77$). This difference is notable but not statistically significant (p-value = 0.11 using two-sample $Z$-test for proportions). The error rate was $7.63 \pm 4.79\%$ for white participants ($n = 118$) and $5.56 \pm 11.39\%$ for non-white participants ($n = 18$). This difference was also non-significant based on Fisher's exact test (Fisher's odd ratio = 0.71, p-value = 1.0). However, the error rate varied notably based on age subgroups. The model performed well for individuals aged 50 to 80. At the same time, the error rate was higher for those aged 30 to 50 or over 80, likely due to the under-representation of these age groups, as the dataset primarily consists of individuals who are 50-80 years old (77.1\% of the entire dataset).

\subsection{Validation with external datasets}
Since we introduced the first multi-task dataset for PD detection in this study, we could not compare our model to similar external datasets. However, we experimented with our model on a recently released public dataset named YouTubePD~\cite{zhou2024youtubepd}. This dataset consists of 283 short YouTube videos collected from 200+ individuals (16 confirmed PD patients). Even without adapting the feature extraction pipeline for this new task, our model (\emph{UFNet} with Smile + Speech modalities) performed comparably to state-of-the-art -- 18\% improvement in accuracy, 3\% reduction in AUROC and $F_1$ score (see Table \ref{tab:youtubepd}). This result is promising as our video features are designed for smile and pangram utterance tasks, vastly different from the videos in the YouTubePD dataset. Adapting the features for this new task should further improve performance.

\setlength{\tabcolsep}{1mm}
\begin{table}[]
\centering
\small
\begin{tabular}{lccc}
\hline
\textbf{Model}                & \textbf{Accuracy}                                                                    & $\mathbf{F_1}$ \textbf{score}                                                                 & \textbf{AUROC}\\ \hline
YouTubePD~\cite{zhou2024youtubepd}
& $70.61\%$          & $\mathbf{61.0\%}$ & $\mathbf{87.0\%}$\\ 

UFNet (Smile + Speech)          & $\mathbf{88.76\%}$          & $58.33\%$          & $83.83\%$\\ \hline
\end{tabular}
\caption{Summary of performance on the YouTubePD dataset. The \textbf{bold} metrics denote best performance.}
\label{tab:youtubepd}
\end{table}

\section{Discussions and Limitations}
Video analysis provides an accessible, cost-effective, and convenient means of screening for PD, particularly benefiting individuals in remote areas or low-income countries with limited access to neurological care. We carefully selected three proposed tasks for remote completion, considering feasibility and safety. While gait analysis is standard for evaluating PD~\cite{li2022detecting, liu2022monitoring}, it presents logistical challenges and potential risks of falling for PD patients. Sustained phonation (holding a vowel sound for as long as possible) is another speech assessment option~\cite{tsanas2009accurate, vaiciukynas2017detecting}, but its analysis becomes complicated due to inconsistencies across various recording devices. All selected tasks demonstrated reasonable predictive performance in differentiating individuals with and without PD, validating our task selection.

Despite our model's promise of accurate PD detection, harm from misprediction remains a concern. Falsely classifying someone as a PD patient may cause undue stress and economic burden, while missing individuals with PD provide a false sense of security and create delays in getting clinical care. With 78.4\% recall, our model may miss some individuals with PD, so we recommend clinical follow-up if symptoms persist or worsen, even if our model does not detect PD. We also advise individuals not to consider our model's positive prediction a definitive diagnosis. Finally, we reduce the harm of mispredictions by withholding predictions when the model is uncertain.

Our model performs consistently across sex and ethnic subgroups, but accuracy drops for younger (30-49 years) and older (over 80 years) age groups. We find that participants from these groups are under-represented, as over 75\% of our dataset participants fall within the 50-80 age range. Until a more balanced dataset is available, we recommend applying the tool only to individuals between 50 and 80. Also, due to requiring an English pangram utterance, our best model cannot be used for non-English speakers. Future work should prioritize recruiting more participants and investigating speech tasks in other languages for the worldwide deployment of our methods.

The decision threshold used in the study could be customized based on individual preferences, as the it directly affects the model's sensitivity and specificity. We used a common 0.5 threshold, but individual preferences for risk-benefit trade-offs might necessitate adjustments. For instance, some users might seek clinical evaluation even at lower probabilities (i.e., prefer high sensitivity), while others might wait for a higher likelihood before incurring healthcare costs (i.e., prefer high specificity).

Furthermore, as videos are primarily gathered unsupervised, issues such as noncompliance with task instructions and various forms of noise are common. For example, during the finger-tapping task, many participants performed fewer than the required ten taps, often with their task-performing hand obscured from view. Background noise may distort speech features, and the presence of multiple individuals in the frame could compromise smile feature extraction. Integrating post hoc quality assessment algorithms into the data collection framework could enhance data quality by identifying quality issues and prompting users to re-record tasks if needed.

A comprehensive, large-scale dataset for Parkinson's disease is currently unavailable due to the lack of resources and clinical collaboration. This paper addresses this gap by introducing the largest multi-task video dataset in the literature and publicly releasing the anonymized features. Unfortunately, we could not gather multiple datasets of videos or extract similar features to compare model performance across different datasets since patient videos (protected health information) are not publicly accessible. Future research on privacy-preserving ways of sharing video datasets can help bridge this gap, facilitating broader comparisons and furthering research in health AI.

\section{Conclusion}
This study demonstrates the promising efficacy of machine learning models in distinguishing individuals with PD from those without PD, requiring only a computer equipped with a webcam, microphone, and internet connection. Given the shared characteristics and nuanced distinctions among movement disorders such as Parkinson's disease, Huntington's disease, ataxia, and Progressive Supranuclear Palsy, these findings hold significant implications. We hope the promising initial results from this research will pave the way for extending tele-neurology applications to encompass a broader range of movement disorders.

\section{Acknowledgments}
The project was supported by the National Science Foundation Award IIS-1750380, National Institute of Neurological Disorders and Stroke of the National Institutes of Health under award number P50NS108676, and Gordon and Betty Moore Foundation. The first author (MSI) is supported by a Google PhD fellowship.

\bibliography{aaai25}

\clearpage

\section{Appendix A. Supplementary Results}

\subsection{Co-variate analysis of demographic variables}
To identify whether there is any sampling bias in our dataset splits, we investigate the distribution of the various demographic attributes across the splits. We found that individuals from different demographic groups are identically distributed across training, validation, and test splits (see Table \ref{tab:covariate}).

\begin{table*}[t]
\begin{tabular}{llllll}
\multicolumn{1}{c}{}                                                 & \multicolumn{1}{c}{}                                   &                                           & \multicolumn{3}{c}{\textbf{Number of Individuals from the Minority Class (\%)}}                    \\
\multicolumn{1}{c}{\multirow{-2}{*}{\textbf{Demographic Attribute}}} & \multicolumn{1}{c}{\multirow{-2}{*}{\textbf{Classes}}} & \multirow{-2}{*}{\textbf{Minority Class}} & \multicolumn{1}{c}{Train (n = 516)} & \multicolumn{1}{c}{Validation (n = 167)} & \multicolumn{1}{c}{Test (n = 162)} \\ \hline
Health Condition                                                     & {with PD, without PD}             & with PD                                   & 168 (32.6\%)               & 55 (32.9\%)                     & 49 (30.3\%)               \\
Sex*                                                                 & Male, Female                                           & Male                                      & 243 (47.1\%)               & 80 (47.1\%)                     & 77 (47.5\%)               \\
Age                                                                  & Below 50, Above 50                                     & Below 50                                  & 48 (9.3\%)                 & 14 (8.4\%)                      & 17 (10.5\%)               \\
Ethnicity                                                            & white, Non-white                                       & Non-white                                 & 44 (8.5\%)                 & 13 (7.8\%)                      & 18 (11.1\%)               \\
Recording Environment                                                & Home, Non-home                                         & Non-home                                  & 193 (37.4\%)               & 72 (43.1\%)                     & 54 (33.3\%)               \\ \hline
\end{tabular}
\caption{\textbf{Co-variate analysis.} Non-binary sex group is not used for simplicity of presentation.}
\label{tab:covariate}
\end{table*}

\subsection{Performance of hand-separated finger-tapping models}
Individuals with PD may experience asymmetry in their motor performance at the onset of the disease. Therefore, to model the finger-tapping task, that assess motor performance, it may be beneficial to utilize both hands' features simultaneously. To investigate this, we trained three task-specific models separately (without employing Monte Carlo dropout). We report their performance in Table \ref{tab:finger-performance}, and observe that utilizing features from both hands significantly improves the $F_1$ score of the model, while maintaining non-inferiority across other metrics.

\begin{table}[H]
    \centering
    \resizebox{1.0\columnwidth}{!}{%
    \begin{tabular}{llccccc}
\multicolumn{2}{l}{\textbf{Task}}           & \textbf{Accuracy}                                                           & \begin{tabular}[c]{@{}c@{}}\textbf{Balanced}\\ \textbf{Accuracy}\end{tabular}        & \textbf{$\mathbf{F_1}$ score}                                                                 & \textbf{AUROC}                                                              & \textbf{AUPRC}                                                              \\ \hline
\multicolumn{2}{l}{Finger-tapping} & \multicolumn{1}{l}{}                                               & \multicolumn{1}{l}{}                                               & \multicolumn{1}{l}{}                                                        & \multicolumn{1}{l}{}                                               & \multicolumn{1}{l}{}                                               \\
            & Both hands           & \begin{tabular}[c]{@{}c@{}}72.0\\ {[}71.1, 72.9{]}\end{tabular} & \begin{tabular}[c]{@{}c@{}}69.0\\ {[}68.1, 70.0{]}\end{tabular} & \begin{tabular}[c]{@{}c@{}}\underline{60.2}\\ {[}58.9, 61.5{]}\end{tabular} & \begin{tabular}[c]{@{}c@{}}73.9\\ {[}73.1, 74.8{]}\end{tabular} & \begin{tabular}[c]{@{}c@{}}58.9\\ {[}57.9, 59.8{]}\end{tabular} \\
            & Left hand            & \begin{tabular}[c]{@{}c@{}}64.3\\ {[}62.9, 65.6{]}\end{tabular} & \begin{tabular}[c]{@{}c@{}}62.0\\ {[}60.7, 63.4{]}\end{tabular} & \begin{tabular}[c]{@{}c@{}}52.6\\ {[}50.8, 54.5{]}\end{tabular}          & \begin{tabular}[c]{@{}c@{}}66.9\\ {[}65.3, 68.6{]}\end{tabular} & \begin{tabular}[c]{@{}c@{}}51.8\\ {[}50.2, 53.4{]}\end{tabular} \\
            & Right hand           & \begin{tabular}[c]{@{}c@{}}73.0\\ {[}72.3, 73.7{]}\end{tabular} & \begin{tabular}[c]{@{}c@{}}70.0\\ {[}68.9, 71.1{]}\end{tabular} & \begin{tabular}[c]{@{}c@{}}47.3\\ {[}44.4, 50.2{]}\end{tabular}          & \begin{tabular}[c]{@{}c@{}}73.7\\ {[}72.5, 75.0{]}\end{tabular} & \begin{tabular}[c]{@{}c@{}}58.5\\ {[}57.0, 60.0{]}\end{tabular} \\ \hline
\end{tabular}%
}
    \caption{Performance of the task-specific finger-tapping models. \underline{Underlined} metrics show significantly better performance.}
    \label{tab:finger-performance}
\end{table}

\subsection{Performance of pre-trained video models}
We experimented with features obtained from four pre-trained video models (VideoMAE~\cite{tong2022videomae}, Uniformer~\cite{li2023uniformer}, ViViT~\cite{arnab2021vivit}, and TimeSformer~\cite{bertasius2021space}) for classifying PD based on the finger-tapping and smile tasks, and compared them against the custom features used in this study. The performance on the finger-tapping task is detailed in Table \ref{tab:finger_features_ablation}, and the smile task in Table \ref{tab:smile_features_ablation}.

\begin{table*}[!htb]
\centering
\begin{tabular}{lccccc}
\textbf{Feature Extraction Method} & \textbf{Accuracy} & \textbf{\begin{tabular}[c]{@{}l@{}}Balanced \\ Accuracy\end{tabular}} & \textbf{F1 Score} & \textbf{AUROC} & \textbf{AUPRC} \\ \hline
Custom features                    & $\mathbf{73.1 \pm 0.7}$    & $\mathbf{70.1 \pm 0.7}$                                                        & $\mathbf{61.7 \pm 0.9}$    & $74.9 \pm 0.7$ & $58.1 \pm 0.9$ \\
VideoMAE                           & $70.2 \pm 0.5$    & $66.8 \pm 0.5$                                                        & $57.0 \pm 0.9$    & $75.1 \pm 0.3$ & $60.6 \pm 0.5$ \\
Uniformer                          & $71.9 \pm 0.4$    & $68.3 \pm 0.6$                                                        & $54.5 \pm 1.6$    & $74.3 \pm 0.6$ & $55.6 \pm 0.8$ \\
ViViT                              & $71.6 \pm 0.5$    & $68.0 \pm 0.6$                                                        & $50.8 \pm 1.9$    & $73.7 \pm 0.4$ & $57.5 \pm 0.5$ \\
TimeSformer                        & $72.6 \pm 0.3$    & $68.9 \pm 0.4$                                                        & $55.9 \pm 1.6$    & $\mathbf{77.2 \pm 0.3}$ & $\mathbf{61.3 \pm 0.5}$ \\ \hline
\end{tabular}
\caption{Finger-tapping task performance}
\label{tab:finger_features_ablation}
\end{table*}

\begin{table*}[!htb]
\centering
\begin{tabular}{lccccc}
\textbf{Feature Extraction Method} & \textbf{Accuracy} & \textbf{\begin{tabular}[c]{@{}l@{}}Balanced \\ Accuracy\end{tabular}} & \textbf{$F_1$ Score} & \textbf{AUROC} & \textbf{AUPRC} \\ \hline
Custom features                    & $\mathbf{77.6 \pm 0.2}$    & $\mathbf{74.4 \pm 0.2}$                                                        & $\mathbf{67.5 \pm 0.3}$    & $\mathbf{83.6 \pm 0.1}$ & $\mathbf{65.4 \pm 0.1}$ \\
VideoMAE                           & $73.5 \pm 0.4$    & $69.1 \pm 0.8$                                                        & $43.2 \pm 2.1$    & $67.3 \pm 0.5$ & $52.5 \pm 0.7$ \\
Uniformer                          & $75.7 \pm 0.4$    & $71.1 \pm 0.6$                                                        & $53.8 \pm 0.7$    & $74.9 \pm 0.4$ & $59.9 \pm 0.3$ \\
ViViT                              & $74.5 \pm 0.5$    & $72.2 \pm 1.0$                                                        & $41.6 \pm 3.0$    & $73.9 \pm 0.4$ & $58.3 \pm 0.8$ \\
TimeSformer                        & $73.9 \pm 0.4$    & $69.4 \pm 0.6$                                                        & $49.1 \pm 1.9$    & $75.5 \pm 0.3$ & $58.9 \pm 0.5$ \\ \hline
\end{tabular}
\caption{Smile task performance}
\label{tab:smile_features_ablation}
\end{table*}

\subsection{Choosing decision threshold}
\begin{figure}[h]
    \centering
    \includegraphics[width=\linewidth]{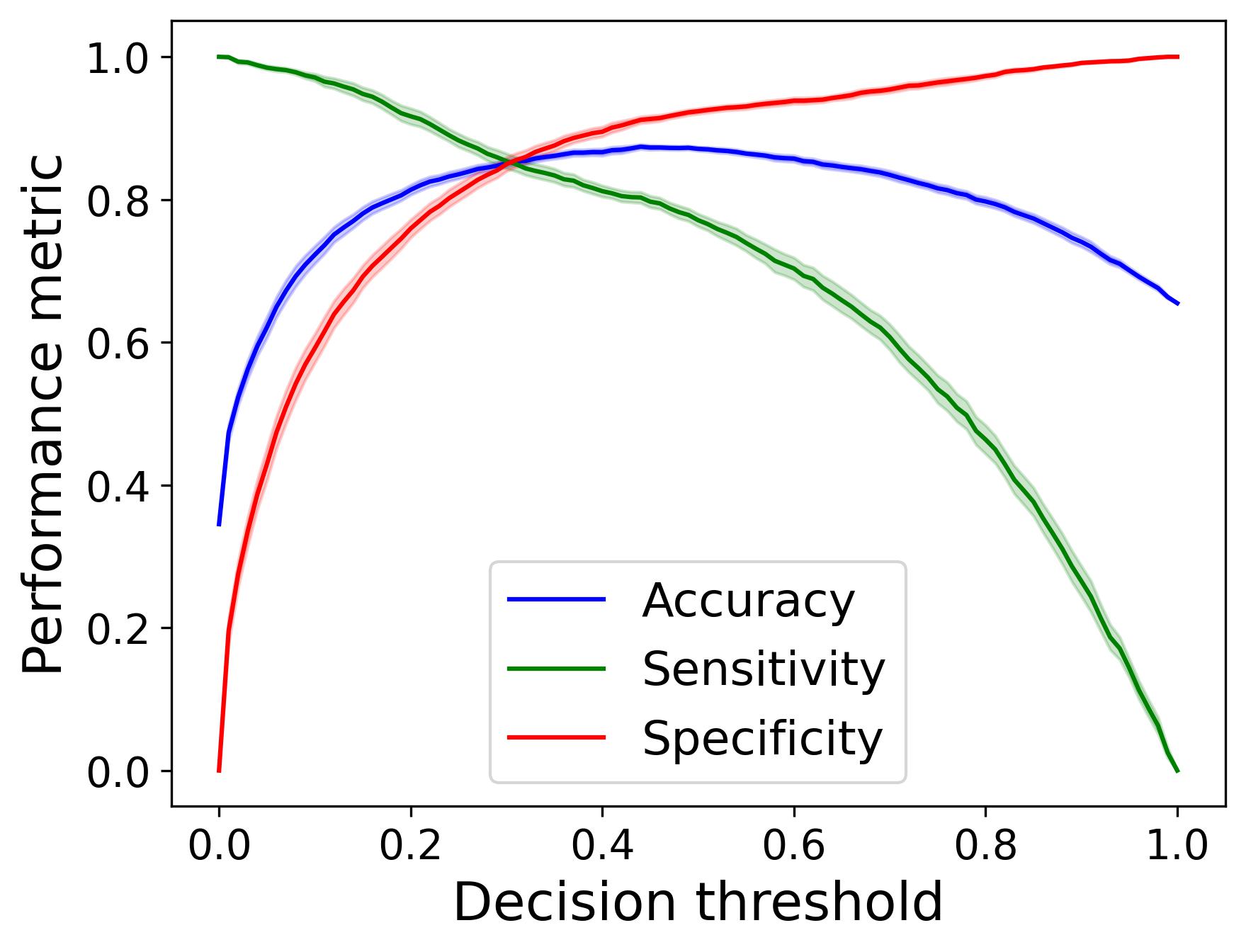}
    \caption{Effect of decision threshold. For this plot, we used the best performing \emph{UFNet} model with 30 different random seeds. Shaded regions show the 95\% confidence intervals.}
    \label{fig:sensitivity_specificity}
\end{figure}
Although it is common practice to choose 0.50 as the threshold for binary classification, this can be further customized based on individual preference and underlying healthcare infrastructure. Here we explore how different decision thresholds impacts the sensitivity and specificity of the proposed \emph{UFNet} model (see Figure \ref{fig:sensitivity_specificity}) on the test set. As expected, with a higher decision threshold, the specificity of the model notably improves, but at the expense of reduced sensitivity. Likewise, a lower decision threshold yields higher sensitivity while the specificity gets penalized. In resource-constrained environments, a higher threshold might be preferred to minimize false negatives. Conversely, when healthcare resources are abundant, a lower threshold can prioritize identifying all potential cases.

\subsection{Performance Analysis across Disease Duration}
To evaluate the model's performance across various stages of PD, we analyzed its misclassification rates relative to disease duration. From 102 participants with duration data, 15 were included in the test set (25 independent samples) covering 10 unique duration values. An initial exploratory analysis using Kendall’s Tau revealed a weak negative correlation between disease duration and misclassification rates (Tau = -0.33, p = 0.39), suggesting a slight trend of decreasing error rates with longer disease duration. However, these findings were not statistically significant due to the small sample size.

\section{Appendix B. Hyper-parameter search for the task-specific models}
\subsection*{Task-specific models without Monte Carlo dropout}

The hyper-parameter search space is outlined in Table \ref{tab:hyp-unimodal-no-mcdrop}. The selected hyper-parameters for the task-specific models are mentioned below:

\begin{table*}
\centering
\begin{tabular}{lll}
\textbf{Hyperparameter}                  & \textbf{Values/range}              & \textbf{Distribution} \\ \hline
batch size                               & \{256, 512, 1024\}                   & Categorical           \\
learning rate                            & [0.0005, 1.0]                      & Uniform               \\
drop correlated features?                & \{``yes'', ``no''\}                      & Categorical           \\
correlation threshold                    & \{0.80, 0.85, 0.90, 0.95\}           & Categorical           \\
use feature scaling?                     & \{``yes'', ``no''\}                      & Categorical           \\
scaling method                           & \{``StandardScaler'', ``MinMaxScaler''\} & Categorical           \\
use minority oversampling (i.e., SMOTE)? & \{``yes'', ``no''\}                      & Categorical           \\
number of hidden layers                  & \{0, 1\}                             & Categorical           \\
number of epochs                         & [1, 100]                           & Uniform Integer       \\
optimizer                                & \{``SGD'', ``AdamW''\}                   & Categorical           \\
momentum                                 & [0.1, 1.0]                         & Uniform               \\
use scheduler?                           & \{``yes'', ``no''\}                      & Categorical           \\
scheduler                                & \{``step'', ``reduce on plateau''\}      & Categorical           \\
step size                                & [1, 30]                            & Uniform Integer       \\
patience                                 & [1, 20]                            & Uniform Integer       \\
gamma                                    & [0.5, 0.95]                        & Uniform               \\
seed
& [100, 999]
& Uniform \\
\hline
\end{tabular}
\caption{Hyper-parameter search space for the task-specific models (without MC dropout).}
\label{tab:hyp-unimodal-no-mcdrop}
\end{table*}

\subsubsection{Finger-tapping task (both hands):} batch size = 256, learning rate = 0.6246956232061768, drop correlated features? = no, use feature scaling? = yes, scaling method = StandardScaler, use minority oversampling? = no, number of hidden layers =  0, number of epochs = 82, optimizer = SGD, momentum = 0.8046223742478498, use scheduler? = no, seed = 276

\subsubsection{Finger-tapping task (left hand):} batch size = 512, learning rate = 0.807750048295928, drop correlated features? = yes, correlation threshold = 0.95, use feature scaling? = yes, scaling method = StandardScaler, use minority oversampling? = no, number of hidden layers =  0, number of epochs = 50, optimizer = SGD, momentum = 0.6614402107331798, use scheduler? = no, seed = 556

\subsubsection{Finger-tapping task (right hand):} batch size = 512, learning rate = 0.5437653223933676, drop correlated features? = no, use feature scaling? = yes, scaling method = StandardScaler, use minority oversampling? = no, number of hidden layers =  1, number of epochs = 74, optimizer = SGD, momentum = 0.709095892070382, use scheduler? = no, seed = 751

\subsubsection{Smile:} batch size = 1024, learning rate = 0.8365099039036598, drop correlated features? = no, use feature scaling? = yes, scaling method = StandardScaler, use minority oversampling? = yes, number of hidden layers =  0, number of epochs = 74, optimizer = SGD, momentum = 0.615229008837764, use scheduler? = yes, scheduler = reduce on plateau, patience = 4, seed = 488

\subsubsection{Speech:} batch size = 256, learning rate = 0.06573643554880117, drop correlated features? = no, use feature scaling? = yes, scaling method = StandardScaler, use minority oversampling? = no, number of hidden layers =  1, number of epochs = 27, optimizer = SGD, momentum = 0.5231696483982686, use scheduler? = no, seed = 287

\subsection*{Task-specific models with Monte Carlo dropout}
The hyper-parameter search space is outlined in Table \ref{tab:hyp-unimodal-with-mcdrop}. The selected hyper-parameters for the task-specific models are mentioned below:

\begin{table*}
\centering
\begin{tabular}{lll}
\textbf{Hyperparameter}                  & \textbf{Values/range}                    & \textbf{Distribution} \\ \hline
batch size                               & \{256, 512, 1024\}                         & Categorical           \\
learning rate                            & [0.0005, 1.0]                            & Uniform               \\
drop correlated features?                & \{``yes'', ``no''\}                            & Categorical           \\
correlation threshold                    & \{0.80, 0.85, 0.90, 0.95\}                 & Categorical           \\
use feature scaling?                     & \{``yes'', ``no''\}                            & Categorical           \\
scaling method                           & \{``StandardScaler'', ``MinMaxScaler''\}       & Categorical           \\
use minority oversampling (i.e., SMOTE)? & \{``yes'', ``no''\}                            & Categorical           \\
number of hidden layers                  & \{0, 1\}                                   & Categorical           \\
dropout probability                      & [0.01, 0.30]                             & Uniform               \\
number MC dropout rounds                 & \{100, 300, 500, 1000, 3000, 5000, 10000\} & Categorical           \\
number of epochs                         & [1, 100]                                 & Uniform Integer       \\
optimizer                                & \{``SGD'', ``AdamW''\}                         & Categorical           \\
momentum                                 & [0.1, 1.0]                               & Uniform               \\
use scheduler?                           & \{``yes'', ``no''\}                            & Categorical           \\
scheduler                                & \{``step'', ``reduce on plateau''\}            & Categorical           \\
step size                                & [1, 30]                                  & Uniform Integer       \\
patience                                 & [1, 20]                                  & Uniform Integer       \\
gamma                                    & [0.5, 0.95]                              & Uniform               \\ \hline
\end{tabular}
\caption{Hyper-parameter search space for the task-specific models (with MC dropout).}
\label{tab:hyp-unimodal-with-mcdrop}
\end{table*}

\subsubsection{Finger-tapping task:} batch size = 256, learning rate = 0.3081959128766984, drop correlated features? = no, use feature scaling? = yes, scaling method = StandardScaler, use minority oversampling? = no, number of hidden layers =  0, dropout probability = 0.24180259124462203, number of MC dropout rounds = 1000, number of epochs = 87, optimizer = SGD, momentum = 0.9206317439937552, use scheduler? = yes, scheduler = reduce on plateau, patience = 13, seed = 790

\subsubsection{Smile task:} batch size = 256, learning rate = 0.03265227174722892, drop correlated features? = no, use feature scaling? = yes, scaling method = StandardScaler, use minority oversampling? = yes, number of hidden layers =  0, dropout probability = 0.10661756438565197, number of MC dropout rounds = 1000, number of epochs = 64, optimizer = SGD, momentum = 0.5450637936769563, use scheduler? = no, seed = 462

\subsubsection{Speech task:} batch size = 1024, learning rate = 0.364654919080181, drop correlated features? = yes, correlation threshold = 0.95, use feature scaling? = no, use minority oversampling? = no, number of hidden layers =  0, dropout probability = 0.23420212038821583, number of MC dropout rounds = 10000, number of epochs = 74, optimizer = AdamW, use scheduler? = no, seed = 303

\section{Appendix C. Hyper-parameter search for the \emph{UFNet} models}

The hyper-parameter search space is outlined in Table \ref{tab:hyp-ufnet}. The selected hyper-parameters are mentioned below:

\begin{table*}[]
\centering
\begin{tabular}{lll}
\textbf{Hyperparameter}           & \textbf{Values/range}                                                                                                          & \textbf{Distribution} \\ \hline
batch size                        & \{$256, 512, 1024$\}                                                                                                               & Categorical           \\
learning rate                     & [$5e^{-5}, 1.0$]                                                                                                                 & Uniform               \\
use minority oversampling?        & \{``yes'', ``no''\}                                                                                                                  & Categorical           \\
oversampling method               & \begin{tabular}[c]{@{}l@{}} \{``SMOTE'', ``SVMSMOTE'', ``ADASYN'',\\ ``BoarderlineSMOTE'', ``SMOTEN'',\\ ``RandomOversampler''\}\end{tabular} & Categorical           \\
number of hidden layers           & \{$1$\}                                                                                                                            & Categorical           \\
projection dimension              & \{$128, 256, 512$\}                                                                                                                & Categorical           \\
query (query/key/value) dimension & \{$32, 64, 128, 256$\}                                                                                                             & Categorical           \\
hidden dimension                  & \{$4, 8, 16, 32, 64, 128$\}                                                                                                        & Categorical           \\
dropout probability               & [$0.05, 0.50$]                                                                                                                   & Uniform               \\
$\eta$                              & [$0.1, 100$]                                                                                                                     & Uniform               \\
number MC dropout rounds          & \{$30$\}                                                                                                                           & Categorical           \\
number of epochs                  & [$1, 300$]                                                                                                                       & Uniform Integer       \\
optimizer                         & \{``SGD'', ``AdamW'', ``RMSprop''\}                                                                                                    & Categorical           \\
momentum                          & [$0.1, 1.0$]                                                                                                                     & Uniform               \\
use scheduler?                    & \{``yes'', ``no''\}                                                                                                                  & Categorical           \\
scheduler                         & \{``step'', ``reduce on plateau''\}                                                                                                  & Categorical           \\
step size                         & [$1, 30$]                                                                                                                        & Uniform Integer       \\
patience                          & [$1, 20$]                                                                                                                        & Uniform Integer       \\
gamma                             & [$0.5, 0.95$]                                                                                                                    & Uniform               \\ \hline
\end{tabular}
\caption{Hyper-parameter search space for the UFNet models.}
\label{tab:hyp-ufnet}
\end{table*}

\noindent \textbf{Finger-tapping + Smile + Speech: } batch size = 1024, learning rate = 0.020724, use minority oversampling? = no, number of hidden layers = 1, projection dimension = 512, query dimension = 64, hidden dimension = 128, dropout probability = 0.4959892, $\eta$ = 81.8179035, number of MC dropout rounds = 30, number of epochs = 164, optimizer = SGD, momentum = 0.689782158, use scheduler? = no, seed=242

\noindent \textbf{Finger-tapping + Smile: } batch size = 256, learning rate = 0.06754950185131235, use minority oversampling? = no, number of hidden layers = 1, projection dimension = 512, query dimension = 64, hidden dimension = 64, dropout probability = 0.4453733432524283, $\eta$ = 12.554916213821272, number of MC dropout rounds = 30, number of epochs = 18, optimizer = SGD, momentum = 0.9822830376765904, use scheduler? = yes, scheduler = reduce on plateau, patience = 10, seed=919

\noindent \textbf{Finger-tapping + Speech: } batch size = 512, learning rate = 0.04035092571261426, use minority oversampling? = no, number of hidden layers = 1, projection dimension = 256, query dimension = 256, hidden dimension = 16, dropout probability = 0.49813214914563847, $\eta$ = 79.95872101951133, number of MC dropout rounds = 30, number of epochs = 164, optimizer = SGD, momentum = 0.24020164138826405, use scheduler? = yes, scheduler = reduce on plateau, patience = 12, seed=953

\noindent \textbf{Smile + Speech: } batch size = 512, learning rate = 0.16688970966723005, use minority oversampling? = no, number of hidden layers = 1, projection dimension = 128, query dimension = 64, hidden dimension = 4, dropout probability = 0.3763157755397192, $\eta$ = 51.88439832518041, number of MC dropout rounds = 30, number of epochs = 132, optimizer = SGD, momentum = 0.22419387711544064, use scheduler? = yes, scheduler = reduce on plateau, patience = 13, seed=845

\vfill\null

\section{Reproducibility Checklist}
\begin{enumerate}
    \item This paper:
        \begin{itemize}
            \item Includes a conceptual outline and/or pseudocode description of AI methods introduced -- yes
            \item Clearly delineates statements that are opinions, hypothesis, and speculation from objective facts and results -- yes
            \item Provides well marked pedagogical references for less-familiare readers to gain background necessary to replicate the paper -- yes
        \end{itemize}
        
    \item Does this paper make theoretical contributions? -- no

    \item Does this paper rely on one or more datasets? -- yes
    \item If yes, please complete the list below.
    \begin{itemize}
        \item A motivation is given for why the experiments are conducted on the selected datasets -- yes
        \item All novel datasets introduced in this paper are included in a data appendix -- partial (cannot share raw video data containing identifiable patient information)
        \item All novel datasets introduced in this paper will be made publicly available upon publication of the paper with a license that allows free usage for research purposes -- partial (cannot share raw video data containing identifiable patient information)
        \item All datasets drawn from the existing literature (potentially including authors’ own previously published work) are accompanied by appropriate citations -- NA
        \item All datasets drawn from the existing literature (potentially including authors’ own previously published work) are publicly available. -- NA
        \item All datasets that are not publicly available are described in detail, with explanation why publicly available alternatives are not scientifically satisficing. -- NA
    \end{itemize}

    \item Does this paper include computational experiments? -- yes
    \item If yes, please complete the list below.
    \begin{itemize}
        \item Any code required for pre-processing data is included in the appendix -- yes
        \item All source code required for conducting and analyzing the experiments is included in a code appendix. -- yes
        \item All source code required for conducting and analyzing the experiments will be made publicly available upon publication of the paper with a license that allows free usage for research purposes. -- yes
        \item If an algorithm depends on randomness, then the method used for setting seeds is described in a way sufficient to allow replication of results. -- yes
        \item This paper specifies the computing infrastructure used for running experiments (hardware and software), including GPU/CPU models; amount of memory; operating system; names and versions of relevant software libraries and frameworks. -- yes
        \item This paper formally describes evaluation metrics used and explains the motivation for choosing these metrics. -- no (evaluation metrics are widely used)
        \item This paper states the number of algorithm runs used to compute each reported result. -- yes
        \item Analysis of experiments goes beyond single-dimensional summaries of performance (e.g., average; median) to include measures of variation, confidence, or other distributional information. -- yes
        \item The significance of any improvement or decrease in performance is judged using appropriate statistical tests (e.g., Wilcoxon signed-rank). -- yes
        \item This paper lists all final (hyper-)parameters used for each model/algorithm in the paper’s experiments. -- yes (in the extended version)
        \item This paper states the number and range of values tried per (hyper-) parameter during development of the paper, along with the criterion used for selecting the final parameter setting. -- yes (in the extended version)
        
    \end{itemize}

\end{enumerate}

\end{document}